\documentclass{article}

\usepackage{arxiv}
\usepackage[utf8]{inputenc} 
\usepackage[T1]{fontenc}    
\usepackage{hyperref}       
\usepackage{url}            
\usepackage{booktabs}       
\usepackage{amsfonts}       
\usepackage[symbol]{footmisc}
\usepackage{nicefrac}       
\usepackage{microtype}      
\usepackage{lipsum}		
\usepackage{graphicx}
\usepackage{doi}
\usepackage{listings}
\usepackage{pythonhighlights}
\usepackage{booktabs}
\usepackage{multicol}
\usepackage[most]{tcolorbox}
\newtcolorbox{entetebox}[1]{
  enhanced,
  colframe=blue!50!black,
  colback=green!10!white,
  colbacktitle=green!10!white,
  fonttitle=\upshape\bfseries,
  fontupper=\itshape,
  drop fuzzy shadow=green!50!black!50!white,
  title={\textcolor{red!50!black}{#1}},
  boxrule=0.4pt
}
\usepackage[superscript]{cite}

\title{Towards a Measure of Algorithm Similarity}

\author{ \href{https://orcid.org/0000-0000-0000-0000}{\includegraphics[scale=0.06]{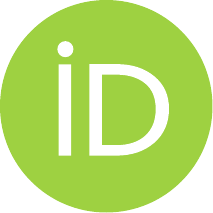}\hspace{1mm}Shairoz Sohail} \\
	\texttt{ShairozSohail@gmail.com} \\
	\And
	\href{https://orcid.org/0000-0000-0000-0000}{\includegraphics[scale=0.06]{orcid.pdf}\hspace{1mm}Taher Ali} \\
	\texttt{taher.dasten@gmail.com} \\
}

\renewcommand{\shorttitle}{\textit{arXiv} Template}

\hypersetup{
pdftitle={A template for the arxiv style},
pdfsubject={q-bio.NC, q-bio.QM},
pdfauthor={David S.~Hippocampus, Elias D.~Striatum},
pdfkeywords={First keyword, Second keyword, More},
}

\begin{document}
\maketitle

\begin{abstract}
	Given two algorithms for the same problem, can we determine whether they are meaningfully different? In full generality, the question is uncomputable, and empirically it is muddied by competing notions of similarity. Yet, in many applications (e.g., clone detection, program synthesis) a pragmatic, consistent similarity metric is necessary. We review existing equivalence and similarity notions and introduce the \emph{EMOC}---an Evaluation-Memory-Operations-Complexity framework that embeds algorithm implementations into a feature space suitable for downstream tasks. We compile PACD, a curated dataset of verified Python implementations across three problems, and show that EMOC features support clustering and classification of algorithm types, detection of near-duplicates, and quantification of diversity in LLM-generated programs. Code, data, and utilities for computing EMOC embeddings are released to facilitate reproducibility and future work on algorithm similarity.
\end{abstract}


\section{Introduction}
The analysis of algorithms has a long and storied history, one that predates any notions of computation \cite{AlgorithmHistory}. At its core, algorithm analysis is about understanding problem-solving procedures. Given a problem and an algorithm to solve it, one can ask many interesting questions:

\begin{enumerate}
    \item Does this algorithm always provide a solution (does it ever stop running)?
  \item Does this algorithm always provide the correct solution? 
  \item What kinds of problem instances is this algorithm well suited for?
  \item What structure of the problem does this algorithm exploit?
  \item As the problem gets more complicated, how much longer does this algorithm take to run?
  \item Is this algorithm the same as a previous algorithm for the problem?\newline
\end{enumerate}

Unfortunately, under the current theoretical frameworks, none of these questions can be answered for an arbitrary algorithm using a computing procedure (that is, they are uncomputable for an arbitrary Turing machine \cite{Sipser2013, RiceTheorem}). Obviously that has neither stopped people from writing algorithms nor people from analyzing them \cite{cormen2009introduction}. The key is that if we meaningfully constrain the space of algorithms and a relevant domain of inputs, carrying out these kinds of analysis is perfectly valid, and at the heart of many fields including complexity theory, compiler design, software validation, and runtime analysis \cite{myers2004art, cooper2004engineering}. 

This raises the question - what kind of constraints can we place on algorithms such that they can be meaningfully compared? The most obvious one is that they solve the same problem, using the same inputs. There is no meaningful comparison between a recipe for baking a cake and the recipe for rebuilding a transmission. It is also dubious to compare a recipe for baking a cake from scratch with one from a mix. However, it is intuitive to compare two recipes for baking a cake that both use the same ingredients - we may ask which one takes longer, which one has more steps, which one results in a tastier cake etc. Similarly, in many fields like Group Theory, we can ask and answer many questions about functions with specific domains (such as the group of isomorphic functions that map from a set to itself \cite{Getz2024}). Thus, we only consider algorithms that have the same domain and attempt to solve the same problem.

Additionally, we would really like to compare algorithms not only as they are written (static analysis) but also as they are executed (dynamic analysis). This requires that the algorithms provide an answer in some reasonable amount of time - otherwise their outputs cannot be compared. Thus, we only consider algorithms that halt on our (pre-selected and finite) domain within a fixed amount of clock time. 

This involves us assuming the answers to question (1) in the positive so we can continue towards answering (6) - when are two algorithms equal? 

\section{Intuition and Formal Definitions}
Utilizing the definition of algorithm as a "set of rules that precisely defines a finite sequence of operations" \cite{stone_definition}, we first observe that algorithms can exist at various levels of abstraction (i.e pseudocode or Turing machine specification). We first differentiate between an algorithm and its implementation. Fixing a language $L$, we write $L(A)$ to illustrate algorithm $A$ implemented in language $L$. It is important to note that a single algorithm $A$ has a potentially infinite number of expressions under a fixed language $L$ (including a minimal one $L^*(A)$, whose derivation may be uncomputable \cite{WOODS2009443}). It is possible to convert an algorithm between two languages $L_1(A) \to L_2(A)$. This is discussed further in section 5.

To build intuitive understanding, we appeal to the area of image classification. Specifically, we consider the mapping from algorithms to algorithm implementations to be the same as the mapping from natural objects to images. Similar to algorithm implementations, this mapping is not one-to-one (there may be many distinct images of a particular object), and there are an infinite number of ways to generate an image of an object by enumerating all possible imaging conditions. 
Importantly, it is common practice to define metrics to measure the separation of two images of the same object and classification procedures to separate them \cite{img_classification}. In this way, we hypothesize that a metric defined on a set of algorithm implementations can serve to reliably discriminate more abstract notions of algorithm equivalency (such as developer intent) as well. 

Formally, Let $P : \{0,1\}^n\mapsto  \{0,1\}^m$ be a computable function that maps from binary\footnote[2]{this may be relaxed to $\mathbb{Z}$ without loss of generality} $n$-dimensional vectors to binary $m$-dimensional vectors, where $m$ may equal $n$ and $P$ may not be bijective. Let $A_1, A_2$ be distinct algorithms that attempt to compute $P$ over a finite domain set $Q = \{q_1... q_n\ | q_i \in \{0,1\}^n\}$. Now, fix a language L \footnote[3]{Note that if $L(A_1), L(A_2)$ are Turing Machines, then under traditional notions of TM equality they are already equal since they compute the same function (over $Q$).In this paper we refer to this as \textbf{functional equality} and try to also take into account the internal logic of $L(A_1), L(A_2)$ in our distance measure.} and consider the (possibly infinite) set of implementations of $A_1, A_2$ under $L$: $\{L^1(A_1), L^2(A_1)...\}$ and  $\{L^1(A_2),L^2(A_2)...\}$ . We would like to define a distance metric $D$ such that:

$$ D(L^i(A_1), L^j(A_1)) < D(L^u(A_1), L^v(A_2)) $$

This constraint forces the distance between implementations of the same algorithm to be less than when compared to an implementation of another algorithm. This ensures invariance to superficial implementation details (such as variable naming or order of associative operations). An additional constraint that would yield even more discriminative power would be:

$$ m = n, i\neq j \Rightarrow D(L^i(A_m), L^j(A_n)) = 0$$

which would force all implementations of the same algorithm to return a zero distance. However, in image embedding such a constraint has been known to cause undesirable properties (such as representation collapse in self-supervised learning \cite{selfsupervised}). We resort to only designing a distance metric that captures the first constraint, in addition to the necessary axioms of distance measures \cite{distance_measures}.

\section{Existing Measures of Similarity}

There are a number of existing notions of algorithm similarity, each with their own advantages and disadvantages. We enumerate the most common approaches below, then categorize them in section 4.6 by modification invariances. 

\subsection{Functional Equivalency}
We consider two Turing Machines $M_1, M_2$ that are complete over some input set $N$. If, for every $n \in N$ we have $M_1(n) = M_2(n)$ (they provide the same output symbol $m$ when started on a tape containing symbol $n$) we say that $M_1 = M_2$. This is known as functional (or extensional) equivalence for computer programs. In general, it is impossible to algorithmically determine functional equivalence for arbitrary machines or programs when the set $N$ is infinite \cite{halting}. However, when the set $N$ is finite, we can determine equivalence by simply comparing the outputs for each $n \in N$. There are an infinite number of algorithms which are functionally equivalent to any given algorithm over a finite set $N$, hence this is the weakest form of algorithm equivalency. 

More broadly (and for algorithms implemented outside of computer languages), functional equivalency says two algorithms are equal when they always produce the same output given the same input. 

\subsection{Instruction Equivalency}
When implemented in a high-level programming language such as Java or Python, an algorithm is often compiled into an intermediate state known as bytecode \cite{thain2016introduction}. Bytecode is not pure machine code and thus has the benefit of being machine independent, often intended for use in a VM (Virtual Machine) such as the JVM (Java Virtual Machine) \cite{jvm} or Python interpreter. When two algorithms compile to identical bytecode, this means that they are functionally equivalent when conditioned on the virtual machine. However, the converse is not true: If two algorithms are functionally equivalent, this says nothing about their potential instruction-bytecode equivalence. Consider the two algorithms below:

\begin{python}
def return181(num):
    solutions = []
    for n in range(1e4):
        for x in range(1e4):
            if (2**n) - 7 == x**2:
                solutions.append(x)
    return(max(solutions))

def return181(num):
    return(181)

\end{python}

While these two algorithms will produce the same answer (181) for any input value, they clearly have a different set of instructions and will compile to different bytecode. Instruction equivalence (as represented by bytecode) is a stronger condition than functional equivalence. 

For algorithms implemented outside of computer languages, it is possible to view the bytecode as simply the set of instructions needed to carry out the algorithm. Clearly it is possible, lacking a complete set of instructions, to arrive at the same set of input/output pairs without reproducing the original set of instructions (an example of functional equivalence without instruction equivalence).  

\subsection{AST equivalency}
An Abstract Syntax Tree (AST) is a higher level representation of a program or algorithm that exists for computer languages which are eventually compiled into bytecode \cite{thain2016introduction}. The purpose of an AST is to provide a human readable representation of the structure of the algorithm. Similar to the relationship between instruction and functional equivalency, AST equivalency guarantees instruction equivalency without the converse being true. Consider the two algorithms below:
\newpage
\begin{python}
def return181_1():
    assert 0==0
    return(181)

def return181_2():
    return(181)

\end{python}

When compiled with the built-in Python compiler, both algorithms are functionally equivalent and instruction equivalent, however they are not AST equivalent. 

For algorithms implemented outside of computer languages, there is no direct analogue of AST equivalence.

\subsection{Encoding/string equivalency}

Finally, we view algorithms as being implemented using a set of symbols (such as the English alphabet or binary number system) and assign equivalency to two algorithms when their encoded representations are the same. This is the strongest and most restrictive form of equivalency - encoding equivalency implies AST equivalency (for computer language encodings), instruction equivalency, and functional equivalency. However, the restrictive nature of encoding equivalency often causes many false negatives (such as those caused by changing the name of an intermediate variable). For any algorithm, there may be (depending on the encoding scheme) an infinite number of ways to break encoding equivalency while maintaining instructional and functional equivalence, but only one way to maintain it - duplicating the algorithm symbol by symbol. 

\section{Difficulties}

Despite the algorithm constraints outlined in section 1 and the notions of equivalency in section 3, there are still numerous challenges to comparing any two algorithms. These informally come down to the fact that while the set of meaningful changes to an algorithm may be small, the set of extraneous or superficial changes is usually infinite. To build intuition, we introduce several examples of "extraneous" changes, how these affect the notions of equivalency described above, and techniques on how to detect and mitigate such changes.

Note that pre-processing algorithm implementations to filter unmeaningful changes (such as those described above) can be seen as trying to approximate the minimal implementation $L^*(A)$ described in section 1.

\subsection{Renaming Variables}
The simplest example of an extraneous change to an algorithm is changing the names of variables. In algorithms implemented in computer languages, such changes are easy to detect and mitigate (by using the AST representation, for example). While this changes the encoding equivalence of two algorithms, it does not change their AST, instruction, or functional equivalence.

\subsection{Extraneous Calculation}
Occasionally, algorithms may contain calculations that are discovered to be non-essential to the final outcome (or redundant to an existing calculation). For example, suppose we are writing a program to check if 3 numbers form a Pythagorean Triple, that is $a^2 + b^2 = c^2$. Our algorithm takes in three positive integers (already sorted, so we do not require a check on which element is c) and returns 1(True) or 0(False). That is our program maps from $\mathbb{N}^3$ to $\{0,1\}$ (commonly called a decision problem \cite{poler2025operations}). One implementation might be as follows:

\begin{python}
def ptriple(a,b,c): 
    return(a**2 + b**2 == c**2)
\end{python}

Now we add an extraneous calculation to the implementation (summing the factorials of $a$ and $b$:

\begin{python}
def ptriple(a,b,c): 
    z = factorial(a) + factorial(b)
    return(a**2 + b**2 == c**2)
\end{python}

The resulting function will still return the same answer on all inputs, however it is meaningfully different in that it calculates an additional quantity (one that happens to take quite a bit of time and memory). Without awareness that the quantity calculated is extraneous and does not affect the output, we could be led to conclude that this is a new algorithm. To avoid this kind of scenario, we must invoke dead code elimination methods \cite{wikipedianscompiler} to compare only the portions of the algorithm that contribute to the result.

\subsection{Intermediate Variables}
Consider again the Pythagorean triple algorithm from above, except with the calculation being split using an intermediate variable $z$

\begin{python}
def ptriple(a,b,c): 
    z = a**2 + b**2
    return(z == c**2)
\end{python}

From a runtime analysis standpoint, this function will consume more memory and compute despite there being no new program logic introduced. Most functions can be arbitrarily split into intermediate variables that are stored to memory instead of computed on the fly, and this is usually a design decision from the programmer. Introduction of an intermediate variable will change the encoding, AST, and instructional (depending on compiler) equivalences of an algorithm, without changing the functional equivalences.

To force invariance to this type of decision,  inlining of intermediate variable assignments can be performed \cite{wikipedianscompiler} before calculating the EMOC score.  

\subsection{Order of Associative Operations}

A more difficult way that two algorithms can be equal is if they implement the same set of associative operations in a different order. Consider the following two implementations of $ptriple$:

\begin{python}
def ptriple1(a,b,c): 
    result = a**2
    result += b**2
    return(result == z)

def ptriple2(a,b,c): 
    result = b**2
    result += a**2
    return(z == result)
\end{python}

When compiled using the built-in Python compiler, both of these algorithms contain different AST and bytecode representations, though they are functionally equivalent. To resolve when two algorithms differ by the order of associative operations, we must be able to determine which of the used operations are associative on the input data types - a highly challenging task even in limited domains such as mathematics and cooking \cite{demoura2008z3}. 

\subsection{Changing of Variable Precision}
Algorithms that operate on the set of real numbers or rational numbers must do so with a fixed level of precision when the machine they are operating on has a finite amount of computational power. In many computer languages, the level of precision can be set manually (and otherwise defaults to the precision of the machine executing the code). Thus, the same algorithm may produce different results (even while being encoding equivalent) when run on machines with different precision. We specifically assume that the algorithm itself does not contain any instructions on the numerical precision desired.

\subsection{Algorithm Changes vs. Maintained Equivalences}

\begin{table}[h]
\resizebox{\textwidth}{!}{%
\begin{tabular}{@{}lllll@{}}
\toprule
 &
  \textbf{Encoding equivalence} &
  \textbf{AST equivalence} &
  \textbf{Instructional equivalence} &
  \textbf{Functional equivalence} \\ \midrule
\multicolumn{1}{l|}{\textbf{Rename variables}} &
  \multicolumn{1}{l|}{Not maintained} &
  \multicolumn{1}{l|}{Maintained} &
  \multicolumn{1}{l|}{Maintained} &
  \multicolumn{1}{l|}{Maintained} \\ \cmidrule(l){2-5} 
\multicolumn{1}{l|}{\textbf{Add extraneous calculation}} &
  \multicolumn{1}{l|}{Not maintained} &
  \multicolumn{1}{l|}{Not maintained} &
  \multicolumn{1}{l|}{Compiler dependent} &
  \multicolumn{1}{l|}{Maintained} \\ \cmidrule(l){2-5} 
\multicolumn{1}{l|}{\textbf{Store intermediate variables}} &
  \multicolumn{1}{l|}{Not maintained} &
  \multicolumn{1}{l|}{Not maintained} &
  \multicolumn{1}{l|}{Compiler dependent} &
  \multicolumn{1}{l|}{Maintained} \\ \cmidrule(l){2-5} 
\multicolumn{1}{l|}{\textbf{Change order of associative operations}} &
  \multicolumn{1}{l|}{Not maintained} &
  \multicolumn{1}{l|}{Not maintained} &
  \multicolumn{1}{l|}{Not maintained} &
  \multicolumn{1}{l|}{Maintained} \\ \cmidrule(l){2-5} 
\multicolumn{1}{l|}{\textbf{Change variable precision}} &
  \multicolumn{1}{l|}{Maintained} &
  \multicolumn{1}{l|}{Maintained} &
  \multicolumn{1}{l|}{Maintained} &
  \multicolumn{1}{l|}{Not maintained} \\ \bottomrule
\end{tabular}%
}
\end{table}

\subsection{A Note on Complexity equivalences}
Another approach to categorizing algorithms is related to the scaling of dynamic properties as the size of the input grows. This approach is central to the area of Complexity Theory \cite{arora2009computational}, with the most commonly studied properties being algorithm runtime and memory consumption. Utilizing algorithm complexity measures as a way to compare algorithms is nuanced. Complexity classes are broad and often encapsulate many different algorithms for a problem under a certain class. Some problems have lower bounds on the best complexity that can be achieved (such as $O(nlog(n))$ for runtime complexity of sorting algorithms that use element comparison \cite{cormen2009introduction}) and thus most algorithms still utilized in practice reach this bound up to some constant. However, it is unlikely that two implementations of the same algorithm have different runtime or memory complexities. Thus, utilizing complexity classes can serve as a strong sufficiency test (in terms of low false positive rate) for two algorithms being different. 

\section{The Correct Abstraction Scale for Algorithm Classification}
Many formal statements about algorithm existence and decidability are stated in terms of Turing Machines or an equivalently powerful model of computation. In this setting, the formality of an algorithm (a Turing Machine) and the problem it solves (computing a well-defined function) are made precise. 

However, the development and assessment of algorithms is often done using the more abstracted languages of programming or mathematics. Introduction of these abstracted languages comes with the "emergence" of high-level properties (such as backtracking, or dynamic programming) that are not well-defined at the levels of Turing Machine instruction. In fact, many seemingly unique operations (such as division and subtraction) become reduced to the same operations at the level of Turing Machine instruction. Thus, it is important to define a point on the "abstraction scale" at which we wish to discriminate algorithms. Similarly to natural systems, the chosen scale may reveal differing points of similarity, and the properties of algorithms may not be scale-invariant. The transition between these scales is the topic of the area of compiler theory. 

\begin{figure}[h]
\raggedleft
\includegraphics[scale=0.7]{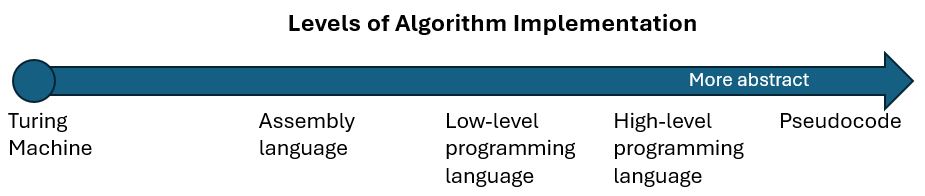}
\caption{Levels of abstraction in algorithm implementation}
\raggedleft
\end{figure}

The runtime analysis included in the EMOC score may minimize the effect of this dependence on implementation level. We believe that the Python programming language provides a sufficiently rich level of abstraction to make static analysis valuable while maintaining enough flexibility to allow for many distinct algorithms (and algorithm design decisions) to be clearly expressed. Discrimination of algorithms in a language-independent manner is an active area of research. \cite{cordy2004practical, vislavski2018licca}

\section{A Dataset for Algorithm Similarity}
To test any algorithm similarity metric, we must first possess a dataset that reflects the natural variation in implementation for different algorithms. This entails a number of requirements:

\begin{enumerate}
\item A variety of algorithmic problems must be considered

\item A variety of algorithms for each problem must be considered

\item All algorithms must be self-contained

\item All algorithms must be in the same programming language

\item The algorithm type must be pre-specified by the developer and then confirmed by a reviewer
\end{enumerate}

To our knowledge, no existing benchmarks serve all of these requirements. Certain clone detection datasets come close, but the elements under consideration are entire scripts and not isolated algorithm implementations, and scripts are not human reviewed for evidence of isolated functionality \cite{zhuo2024bigcodebench, daghighfarsoodeh2025deep, zavershynskyi2018naps} . Thus, we compile and publicly release a small dataset of unconstrained algorithm implementations in Python. 

Our dataset, PACD (Python Algorithm Classification Dataset), consists of 350 algorithms implementations against 3 problems (list sorting, PRIMES, and list search). For each problem,  different algorithms are compiled with manual verification of each algorithm for correctness, accuracy, and lack of runtime bugs.  

\section{EMOC Components, Algorithm Encoding}
\subsection{A Note on Function Domains and Sampling}
One challenge for sampling based analysis of algorithms is that algorithm runtime is not consistent on all cases of a fixed size. Certain arrangements (such as a list in reverse sorted order) can require substantially more time/runtime operations than an "average" case - thus empirical analysis of algorithms may easily be biased by the likelihood that a worst-case example is generated during random sampling \cite{cormen2009introduction}. 

We bypass this by explicitly generating worst-case examples (when they are known for the algorithm) or by enforcing highly varied arrangements using maximal variance sampling strategies. As a companion to the dataset provided above, we provide sample inputs for each algorithm that include worst-case and best-case examples. 

\subsection{E-Component}
The notion of functional equivalency outlined in section 3.1 can be approximated by evaluating algorithms against a carefully sampled set of inputs. When disagreements occur, we can state that (up to the precision of the machine evaluating the algorithms) they are functionally unequal. We assign a value of 1 to the E-component when there is a disagreement in output values for the same input value, otherwise we assign a 0. Finer-grained notions of difficulty rely on manually setting a tolerance unit and are problem-specific.

\subsection{M-Component}

Another notion of algorithm equivalence is memory complexity. As described in section 4.7, differences in complexity measures like scaling of memory usage serve as a necessary but not sufficient condition to declare two algorithms to be the "same". To estimate the M-component, we provide inputs of increasing size to the algorithm and take the ratio of successive inputs to approximate the scaling behavior. For example, with 10 inputs of increasing size we would end up with 9 ratios of memory usage between each pair of inputs. This allows for some invariance to overall memory usage and approximates the pure scaling effect. 

\subsection{O-Component}

To approximate instruction equivalence, we pick a set of primitive operations (such as addition, multiplication, bitwise XOR etc.), Then, two algorithms can be set to be instruction equivalent if they utilize the same operations in the same order. If the set of operations is associative, then we can discard the requirement for them to be in the same order. Given the complex nature of determining circumstances where this associativity holds, we use the much simpler format of simply counting the number of distinct primitive operations and encoding the results into a N-length count vector, where N is the number of primitive operations we would like to consider.  

\subsection{C-Component}

Similar to the M-component, we can utilize the runtime complexity of an algorithm to compare it to other algorithms for the same problem. This term is estimated the same way as the M-component, except utilizing runtime estimates (in clock time) for inputs of increasing sizes instead of memory usage. 

\newpage

\section{Results}

Observe that combining the outputs from the E-component (single binary value), M-component (n-length vector of floats), O-component (k-length vector of binary values), and C-component (m-length vector of floats) allows us to compute a $1+N+K+M$ length numeric vector embedding of an algorithm. This vector represents what we believe to be all the key elements of an algorithm needed to differentiate it from other algorithms for the same task.

\subsection{Classifying Algorithm Implementations using EMOC}
Utilizing our dataset and different components of the EMOC metric, we can set up a natural classification task with the target being algorithm type (as specified by the developer, and vetted by manual curation). 

For this classification task, we utilize K-means clustering with K set to the number of different algorithms present in the dataset for the specific problem. 

The benefit of utilizing a K-means approach is it can be replicated for data where we don't have ground truth labels (such as output from a large language model, as discussed in section 8.3). 

Below are the results for utilizing this procedure to cluster the sorting algorithms in our PACD dataset. While only the C-term and M-term are shown, the full EMOC score was utilizing for the clustering. 

\begin{figure}[h]
    \centering
    \includegraphics[width=1\linewidth]{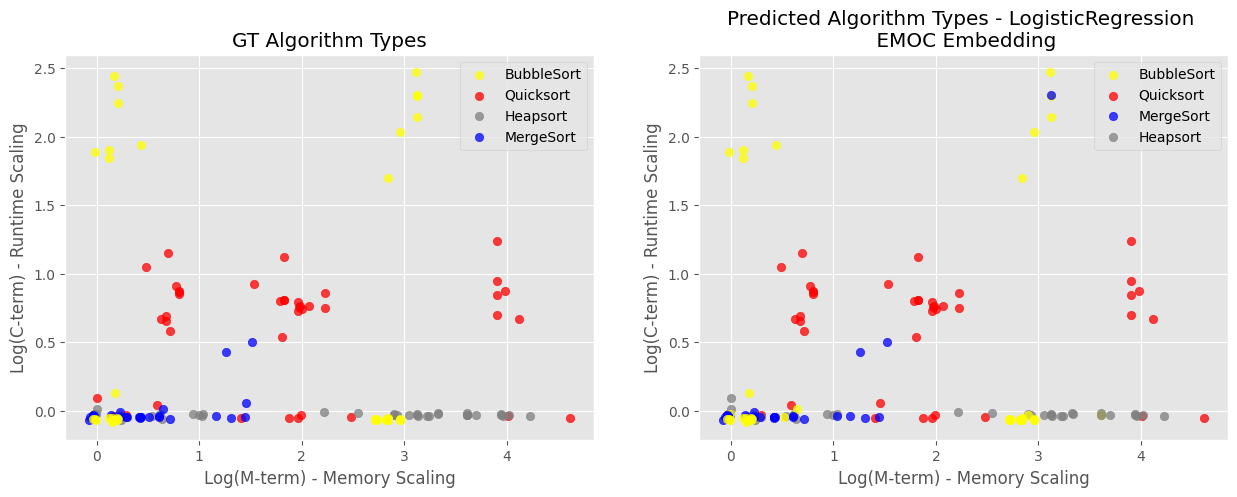}
    \caption{Utilizing EMOC Embeddings to Classify Sorting Algorithm Implementations from PACD (Test acc = 79.1\%)}
    \label{fig:enter-label}
\end{figure}

\subsection{Clone Detection using EMOC}
In computing, a program clone refers to creating a duplicate of a program, often involving replicating the program's settings, activities, and even its source code \cite{cordy2004practical}. We can use the EMOC embedding of an algorithm to find it's nearest neighbors (other algorithm implementations that have the same outputs, similar memory/runtime scaling, and similar operations). This can sometimes yield surprising results. For example, below are the two most similar BubbleSort implementations in PACD:

\begin{python}
def bubbleSort(alist):
    for passnum in range(len(alist)-1,0,-1):
        for i in range(passnum):
            if alist[i]>alist[i+1]:
                temp = alist[i]
                alist[i] = alist[i+1]
                alist[i+1] = temp
    return(alist)
\end{python}

\newpage
\begin{python}
def ordenar_lista(lista):
	# obtém o tamanho da lista
	len_lista = len(lista)
	# loop mais externo começando de (len_lista - 1) até 1 com decremento -1
	for i in range(len_lista - 1, 0, -1):
		# for mais interno que vai de 0 até (i - 1)
		for j in range(i):
			# a função lower() converte todos os caracteres de uma string para minúsculo
			# isso serve para ignorar o case-sensitive
			# o "if" testa se a string acessada pelo índice (j + 1)
			# precede a string acessada pelo índice "j"
			if(lista[j] > lista[j + 1]):
				# faz a troca dos elementos
				aux = lista[j]
				lista[j] = lista[j + 1]
				lista[j + 1] = aux
      
	return(lista)
\end{python}

\subsection{Measuring Novelty in Large Language Model Program Synthesis}

Equipped with a measure that captures algorithm similarity, we attempt to use it to answer several questions about LLM-based algorithm generation: \newline

1) Does the diversity of LLM generated algorithms correlate with sampling temperature?

2) Do LLMs with more parameters generate more diverse algorithm types?

3) Can we directly bias an LLM to generate novel algorithms via the input prompt? \newline

We tackle these questions in the above order. 

For (1), we fix the problem (list sorting) and LLM while increasing the sampling temperature from  0.75 to 1.1. We then inspect the variance in each of the EMOC components for the generated algorithms from each sampling temperature. The E-component has 0 variance, indicating all (functional) sorting algorithms from the LLM (gpt-4o) performed sorting correctly. Below, we plot the differences in the O-component:

Given two algorithms for the same problem, can we determine whether they are meaningfully
different? In full generality, the question is uncomputable, and empirically it is muddied by competing
notions of similarity. Yet, in many applications (such as clone detection and program synthesis) a pragmatic,
consistent similarity metric is necessary. We review existing equivalence and similarity notions
and introduce EMOC: An Evaluation-Memory-Operations-Complexity framework that embeds
algorithm implementations into a feature space suitable for downstream tasks. We compile PACD,
a curated dataset of verified Python implementations across three problems, and show that EMOC
features support clustering and classification of algorithm types, detection of near-duplicates, and
quantification of diversity in LLM-generated programs. Code, data, and utilities for computing
EMOC embeddings are released to facilitate reproducibility and future work on algorithm similarity.

\begin{figure}[h]
    \centering
    \includegraphics[width=1\linewidth]{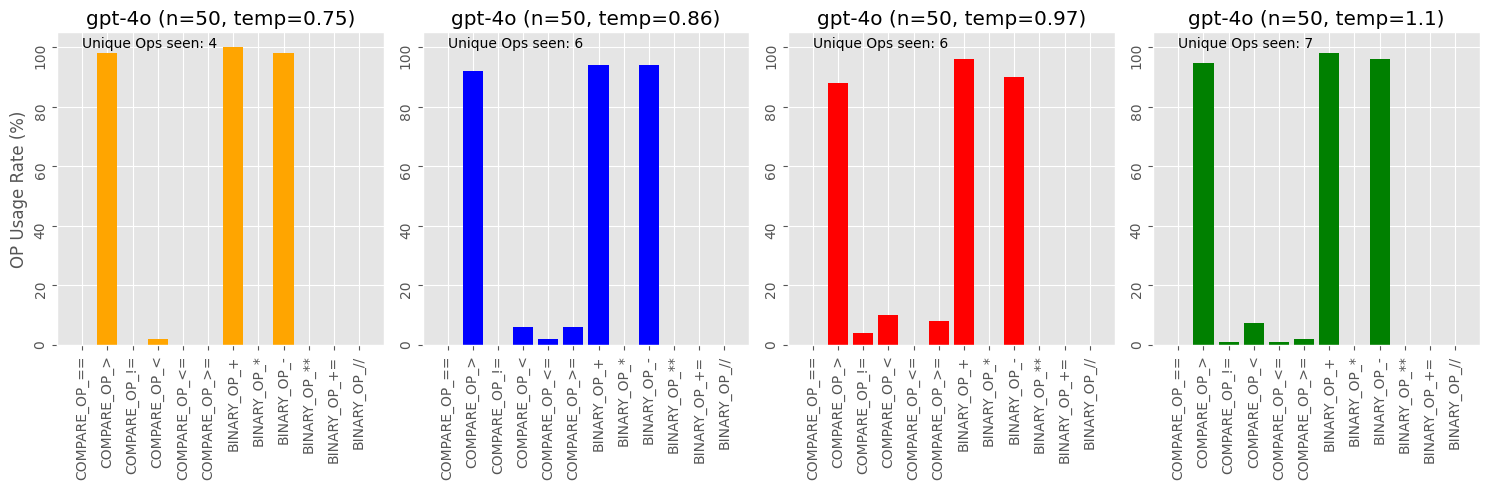}
    \caption{Utilizing EMOC embeddings to measure diversity of LLM generated sorting algorithms}
    \label{fig:enter-label}
\end{figure}

We can observe that as sampling temperature increases, we see more unique operations utilized in the sorting algorithms generated by the LLM. \newline

For (2), we fix the problem (list sorting) and the sampling temperature (1.0), and utilize models with an increasing number of parameters. 

\begin{figure}[h]
    \includegraphics[scale = 0.5]{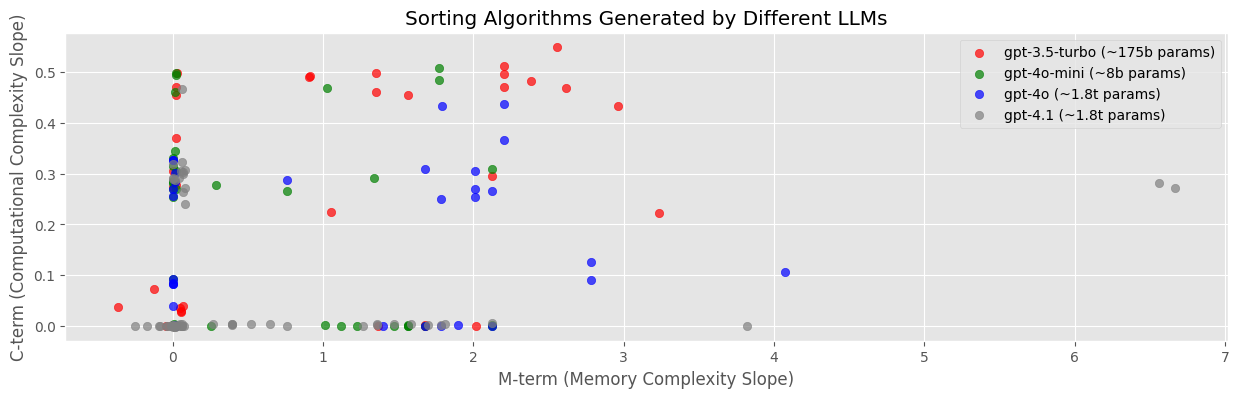}
    \caption{Directly encouraging novelty by modifying prompt}
    \label{fig:enter-label}
\end{figure}

When LLMs are trained with higher parameter counts, the resulting algorithms they generate are more varied (as measured by memory and computational complexity). Furthermore, this trend holds when looking at the primitive operations utilized by the algorithms:

\begin{figure}[h]
    \includegraphics[scale = 0.5]{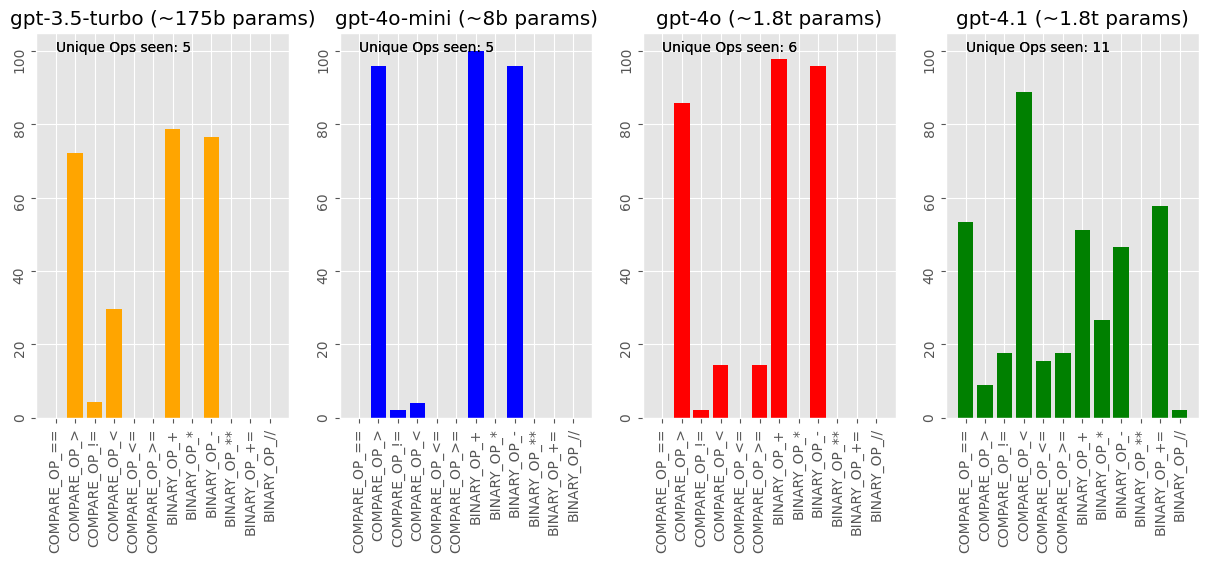}
    \caption{Diversity of generated algorithms by LLM size}
    \label{fig:enter-label}
\end{figure}

For (3), we fix the problem (list sorting), the LLM (gpt-4o), and change the prompt to attempt to directly encourage novelty:

\begin{entetebox}{Original prompt}
  Write a function that takes in a list of integers and returns the list in sorted ascending order. Do not use the builtin sorted() function.
\end{entetebox}

\begin{entetebox}{Modified prompt}
  Write a function that takes in a list of integers and returns the list in sorted ascending order. Do not use the builtin sorted() function. Focus on generating a novel or unique algorithm.
\end{entetebox}

\begin{figure}[h]
    \center
    \includegraphics[scale = 0.45]{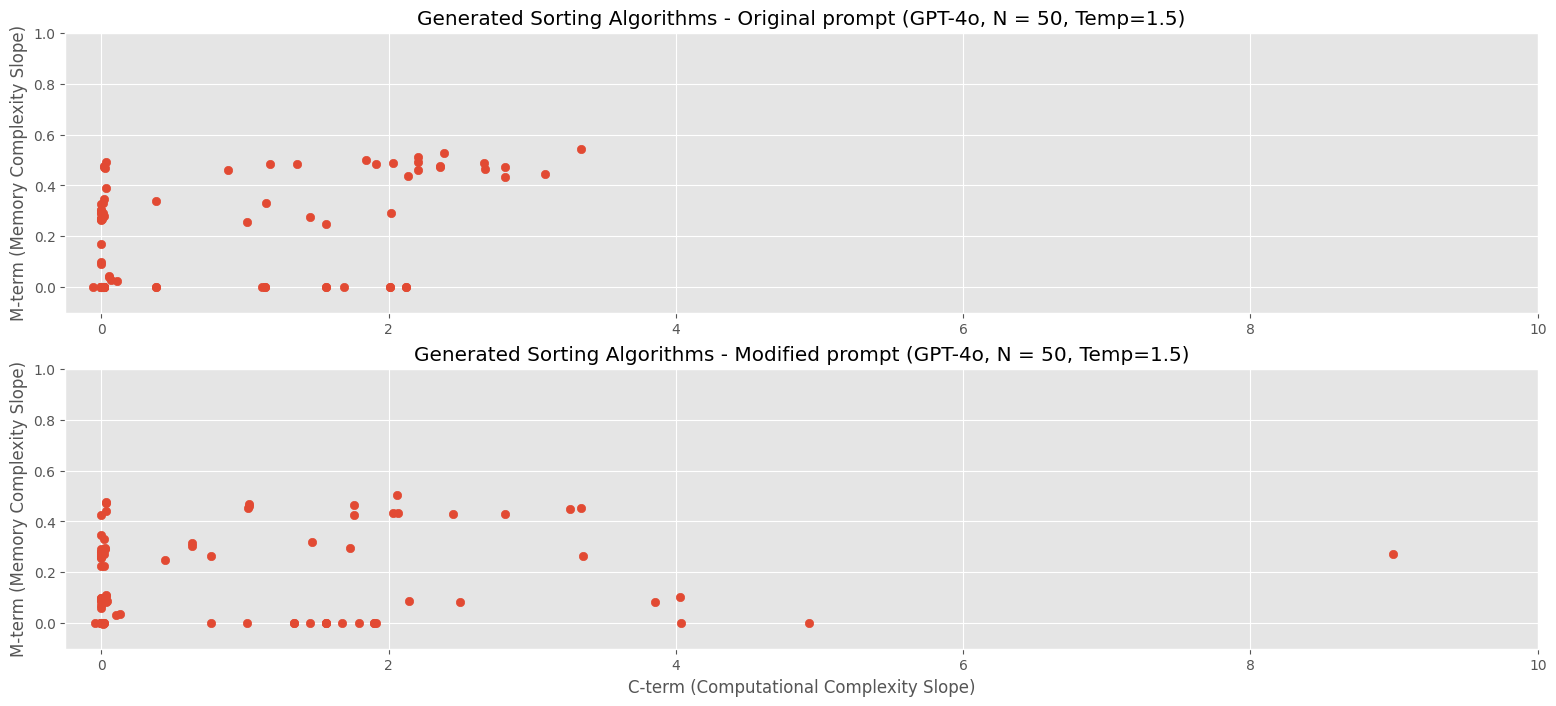}
    \caption{Diversity in generated algorithms by prompt used}
\end{figure}

\newpage
\newpage 

\section{Conclusions and Future Work}
Our results show promising utility for the EMOC criteria as a method for embedding algorithms into a space
where they can be meaningfully compared. We demonstrate the ability to perform classification on
both human and LLM generated sorting algorithms, predicting labels that are consistent with the original intention of the algorithm authors. We also demonstrate the ability to utilize EMOC scores as a way to detect potentially novel algorithms generated by LLM sampling. A natural application is to directly utilize the EMOC score within the LLM prompt. This may provide a more direct method of encouraging novelty as compared to simply increasing the sampling temperature, which did not yield a high variety of algorithm types
in our tests. We hope that this framework (along with the provided codebase) finds utility not only in traditional algorithm
analysis tasks but also in supporting the discovery of novel algorithms using deep generative models.

\bibliography{references}

\begin{thebibliography}{10}

\bibitem{AlgorithmHistory}
{Alan Beadle}.
\newblock Historical origins of data structures and algorithms.

\bibitem{arora2009computational}
Sanjeev Arora and Boaz Barak.
\newblock {\em Computational Complexity: A Modern Approach}.
\newblock Cambridge University Press, Cambridge; New York, 2009.

\bibitem{cooper2004engineering}
K.D. Cooper and L.~Torczon.
\newblock {\em Engineering a Compiler}.
\newblock Elsevier Science, 2004.

\bibitem{cordy2004practical}
James~R. Cordy, J.~Dingel, T.~Wigg, and C.~Dean.
\newblock A practical language-independent approach to detecting cloned code in large software systems.
\newblock In {\em International Conference on Software Maintenance (ICSM)}, pages 156--165. IEEE, 2004.

\bibitem{cormen2009introduction}
Thomas~H Cormen, Charles~E Leiserson, Ronald~L Rivest, and Clifford Stein.
\newblock {\em Introduction to Algorithms}.
\newblock MIT Press, 3rd edition, 2009.

\bibitem{daghighfarsoodeh2025deep}
S.~Daghighfarsoodeh, A.~A. Ghorbani, and Q.~Wang.
\newblock Deep learning for source code vulnerability detection: A systematic review.
\newblock {\em IEEE Transactions on Software Engineering}, 2025.
\newblock Scheduled for publication or accepted. Assuming future date is correct.

\bibitem{demoura2008z3}
Leonardo De~Moura and Nikolaj Bj{\o}rner.
\newblock {Z3}: An efficient {SMT} solver.
\newblock In {\em International Conference on Tools and Algorithms for the Construction and Analysis of Systems}, pages 337--340. Springer, 2008.

\bibitem{Getz2024}
Jayce~R. Getz and Heekyoung Hahn.
\newblock {\em An Introduction to Automorphic Representations: With a view toward trace formulae}, volume 300 of {\em Graduate Texts in Mathematics}.
\newblock Springer International Publishing, 2024.

\bibitem{img_classification}
Alex Krizhevsky.
\newblock Learning multiple layers of features from tiny images.
\newblock Technical report, University of Toronto, 2009.
\newblock \url{http://www.cs.toronto.edu/~kriz/cifar.html}.

\bibitem{jvm}
Tim Lindholm, Frank Yellin, Gilad Bracha, and Alex Buckley.
\newblock {\em The Java Virtual Machine Specification: Java SE 9 Edition}.
\newblock Oracle America, Inc., 2017.
\newblock \url{https://docs.oracle.com/javase/specs/jvms/se9/html/index.html}.

\bibitem{distance_measures}
James~R. Munkres.
\newblock {\em Topology}.
\newblock Prentice Hall, second edition, 2000.

\bibitem{myers2004art}
G.J. Myers, C.~Sandler, T.~Badgett, and T.M. Thomas.
\newblock {\em The Art of Software Testing}.
\newblock Business Data Processing: A Wiley Series. Wiley, 2004.

\bibitem{poler2025operations}
R.~Poler, J.~Mula, M.~D{\'\i}az-Madro{\~n}ero, and R.~Sanchis.
\newblock {\em Operations Research Problems: Statements and Solutions}.
\newblock Springer London, 2025.

\bibitem{Sipser2013}
Michael Sipser.
\newblock {\em Introduction to the Theory of Computation}.
\newblock Cengage Learning, Boston, MA, third edition, 2013.

\bibitem{stone_definition}
Harold~S. Stone.
\newblock {\em Introduction to Computer Organization and Data Structures}.
\newblock McGraw-Hill, New York, 1971.

\bibitem{thain2016introduction}
D.~Thain.
\newblock {\em Introduction to Compilers and Language Design}.
\newblock Lulu Press, Incorporated, 2016.

\bibitem{selfsupervised}
Yuandong Tian, Xinlei Chen, and Surya Ganguli.
\newblock Understanding self-supervised learning dynamics without contrastive pairs.
\newblock In {\em International Conference on Machine Learning}, pages 10268--10278. PMLR, 2021.

\bibitem{halting}
A.~M. Turing.
\newblock On computable numbers, with an application to the entscheidungsproblem.
\newblock {\em Proceedings of the London Mathematical Society}, 2(1):230--265, 1937.

\bibitem{vislavski2018licca}
Yevgen Vislavski and Rik Wettel.
\newblock {LiCCA}: A language-independent code clone analysis approach based on abstract syntax trees.
\newblock In {\em International Working Conference on Source Code Analysis and Manipulation (SCAM)}, pages 21--30. IEEE, 2018.

\bibitem{RiceTheorem}
Mahesh Viswanathan.
\newblock Rice's theorem.

\bibitem{wikipedianscompiler}
B.~Wikipedians.
\newblock {\em Compiler Construction}.
\newblock PediaPress.

\bibitem{WOODS2009443}
Damien Woods and Turlough Neary.
\newblock The complexity of small universal turing machines: A survey.
\newblock {\em Theoretical Computer Science}, 410(4):443--450, 2009.
\newblock Computational Paradigms from Nature.

\bibitem{zavershynskyi2018naps}
Mykola Zavershynskyi, Vladyslav Kurbatov, Svitlana Zhydenko, and Iryna Gurevych.
\newblock {NAPS}: Neural abstractive {P}rogram {S}ummarization.
\newblock In {\em Annual Meeting of the Association for Computational Linguistics (ACL)}, pages 2291--2301, 2018.

\bibitem{zhuo2024bigcodebench}
Siyuan Zhuo, Ziyang Li, Jinbo Huang, and et~al.
\newblock {BigCodeBench}: A comprehensive benchmark for large language models of code.
\newblock {\em arXiv preprint arXiv:2407.03961}, 2024.

\end{thebibliography}

\end{document}